\documentclass[conference]{IEEEtran}
\IEEEoverridecommandlockouts
\usepackage{cite}
\usepackage{amsmath,amssymb,amsfonts}
\usepackage[ruled,vlined,noend]{algorithm2e}
\usepackage{algpseudocode}
\usepackage{graphicx}
\usepackage{textcomp}
\usepackage{multirow}
\usepackage{multicol}
\usepackage{xcolor}

\usepackage{fancyhdr}

\def\BibTeX{{\rm B\kern-.05em{\sc i\kern-.025em b}\kern-.08em
    T\kern-.1667em\lower.7ex\hbox{E}\kern-.125emX}}
\begin{document}

\title{Theoretical Rule-based Knowledge Graph Reasoning by Connectivity Dependency Discovery\\
\thanks{We appreciate the support from IBM research, and the Center of Microbiome Innovation in UC San Diego.

© 2022 IEEE. This paper is accepted at IEEE International Joint Conference on Neural Networks (IJCNN) 2022. This is the preprint version.}
}

\author{\IEEEauthorblockN{1\textsuperscript{st} Canlin Zhang}
\IEEEauthorblockA{\textit{Circulo Health (present)} \\
\textit{Center of Microbiome Innovation} \\
\textit{University of California, San Diego (sponsor)} \\
Columbus, US \\
canlin.zhang@circulohealth.com}
\and
\IEEEauthorblockN{2\textsuperscript{nd} Chun-Nan Hsu}
\IEEEauthorblockA{\textit{$ \ \ \ \ \ \ \ \ $ Center for Research in Biological Systems $ \ \ \ \ \ \ \ \ $} \\
\textit{University of California, San Diego} \\
La Jolla, US \\
chunnan@ucsd.edu}
\and
\IEEEauthorblockN{3\textsuperscript{th} Yannis Katsis}
\IEEEauthorblockA{\textit{IBM Research-Almaden} \\
San Jose, US}
\and
\IEEEauthorblockN{4\textsuperscript{th} Ho-Cheol Kim}
\IEEEauthorblockA{\textit{IBM Research-Almaden} \\
San Jose, US}
\and
\IEEEauthorblockN{5\textsuperscript{th} Yoshiki V\'azquez-Baeza}
\IEEEauthorblockA{\textit{University of California, San Diego} \\
La Jolla, US}}

\maketitle
\thispagestyle{fancy}

\begin{abstract}
Discovering precise and interpretable rules from knowledge graphs is regarded as an essential challenge, which can improve the performances of many downstream tasks and even provide new ways to approach some Natural Language Processing research topics. In this paper, we present a fundamental theory for rule-based knowledge graph reasoning, based on which the connectivity dependencies in the graph are captured via multiple rule types. It is the first time for some of these rule types in a knowledge graph to be considered. Based on these rule types, our theory can provide precise interpretations to unknown triples. Then, we implement our theory by what we call the \emph{RuleDict} model. Results show that our RuleDict model not only provides precise rules to interpret new triples, but also achieves state-of-the-art performances on one benchmark knowledge graph completion task, and is competitive on other tasks. 
\end{abstract}


\section{Introduction}
A knowledge graph (KG) is a graphical representation of the knowledge base (KB), in which entities are represented by nodes and relations are represented by links among nodes. Knowledge graphs are useful tools in many Natural Language Processing (NLP) research areas, such as question answering~\cite{Zhang_KG_QandA,Huang_KG_QandA,Diefenbach_KG_QandA}, semantic parsing~\cite{Yih_KG_semantic_parsing,Berant_semantic_parsing} and dialogue systems~\cite{He_KG_dialogue,Keizer_KG_dialogue}. However, most knowledge graphs suffer from missing relations~\cite{Socher_KG_reasoning,West_KG_KB_completion}, which leads to the task of \emph{knowledge graph completion} or \emph{link prediction}. The task aims at recovering missing relations in a KG given the known ones.

In general, there are two approaches to the task of knowledge graph completion: the Embedding-based approach~\cite{TransE,ConvE,ConvKB} and the rule-based approach~\cite{NeuralLP,MINERVA,M3GM}. Generally speaking, embedding-based models represent entities and relations as real-valued vectors~\cite{mikolov2013distributed}. Then, deep neural networks are trained in an end-to-end manner~\cite{end_to_end_NLP} based on these vectorized embeddings to capture the semantic information in a knowledge graph. 

Although being flexible and expressive~\cite{Socher_KG_reasoning}, embedding-based models cannot provide definite explanations behind their link prediction results~\cite{limits_end_to_end,limits_embeddings}. In contrast, rule-based models, although suffering from scalability issues and lack of expressive power, attempt to capture the inherent regularities in a knowledge graph as rules. Link prediction results from these models can be explained by how the rules are followed~\cite{NeuralLP,MINERVA}. For example, suppose we have the rule: \textit{All monkeys have the part tail} (We just assume the correctness of this rule here). Then, since \textit{Yunnan snub-nosed monkey} is a subspecies (hyponym) of \textit{monkey}, we will know that \textit{Yunnan snub-nosed monkey} has the part \textit{tail}.


In this paper, we present a rule-based knowledge graph completion model using different types of rules, which provides a large amount of rules in a series of types to evaluate the correctness of unknown triples. Our contribution is three-fold: 

(i) We establish a fundamental theory for knowledge graph reasoning, based on which any knowledge graph can be completed in a interpretable way. 

(ii) We discover two novel rule types, which are never considered in previous knowledge graph reasoning models.

(iii) We come up with the \emph{RuleDict} model, an efficient implementation of our theory, which can provide precise rules in different types, explaining why a triple is correct or not. 

In Section~\ref{sec:method}, we introduce our knowledge graph reasoning theory in detail. In Section~\ref{sec:imp_perf}, we describe how to implement our theory by the RuleDict model, and then present our experimental results on large benchmark datasets of knowledge graph completion. Section~\ref{sec:disc} further discusses our experimental results, with an analysis on some of our example outputs. 
Related work is surveyed in Section~\ref{sec:related_work}. Finally, we conclude this paper and provide our future research plans in Section~\ref{sec:conclusion}.

\section{Methodology}
\label{sec:method}
In this section, we will first describe the problem of knowledge graph reasoning as well as the related concepts. Then, we will introduce our knowledge graph reasoning theory.

\subsection{Description on the Problem and Concepts}

A knowledge graph is denoted by $\mathcal{G}= (\mathcal{E},\mathcal{R},\mathcal{T})$,
where $\mathcal{E}$ and $\mathcal{R}$ represent the set of \emph{entities}
(nodes) and \emph{relations} (links), respectively. Then, a \emph{triple} $r(s,t)$ in the known triple set $\mathcal{T}$ indicates that there is a relation $r$ between entities $s$ and $t$. Triples are the atoms of a knowledge graph $\mathcal{G}$. Then, we can also consider that $\mathcal{G}$ is generated by the known triples $\{r_i(s_i,t_i)\}_{i=1}^{N_{\mathcal{G}}}$. 


In most knowledge graphs, the relation $r$ is directed. For instance, \emph{has$\_$part(monkey, tail)} is a known triple in WordNet while \emph{has$\_$part(tail, monkey)} is not~\cite{Socher_KG_reasoning}. Then, the inverse relation $r^{-1}$ is defined for each $r\in \mathcal{R}$, so that $r^{-1}(t,s)$ is a known triple whenever $r(s,t)$ is. Accordingly, the relation set becomes $\mathcal{R}\cup \mathcal{R}^{-1}$ with $\mathcal{R}^{-1}=\{r^{-1}\}$; and the known triple set becomes $\mathcal{T}\cup\mathcal{T}^{-1}$ with $\mathcal{T}^{-1}=\{r_i^{-1}(t_i,s_i)\}_{i=1}^{N_{\mathcal{G}}}$. Hence, the inverse completed knowledge graph will be $\tilde{\mathcal{G}}=(\mathcal{E},\mathcal{R}\cup\mathcal{R}^{-1},\mathcal{T}\cup\mathcal{T}^{-1})$. To minimize confusion, we still use the symbol $\mathcal{G}$.

We use capital-case symbols $X,Y,\ldots$ to denote free entities and use lower-case ones $s,t,\ldots$ to denote anchored entities. We say that there is a \emph{path} (or \emph{Horn clause path}) $r_1\!\land\! r_2\!\land\!\cdots\!\land\!r_l$ between entities $s$ and $t$, if there are entities $s_2,\cdots s_l$ in $\mathcal{E}$ such that $r_1(s,s_2), r_2(s_2,s_3),\cdots, r_l(s_l,t)$ are known triples in $\mathcal{T}\cup\mathcal{T}^{-1}$. Similarly, we claim that an $l$-hop \emph{ending anchored structure} (EAS) $r_1(X_1,X_2)\!\land\! r_2(X_2,X_3)\!\land\!\cdots\!\land\!r_l(X_l,t)$ exists in $\mathcal{G}$, if there are entities $s_1,\cdots s_l$ in $\mathcal{E}$ such that $r_1(s_1,s_2), r_2(s_2,s_3),\cdots, r_l(s_l,t)$ are known triples in $\mathcal{T}\cup\mathcal{T}^{-1}$. 

While taking paths with different lengths into consideration, in this paper, we only focus on the one-hop EAS $r(X,t)$. In the following parts, by ending anchored structure or EAS, we always mean one-hop EAS. 
We will denote an EAS using lower-case letter $a$ or $b$, such as $a=r(X,t)$ or $b=r'(X,t')$. Similar to entities, we may use $a,b\in\mathcal{G}$ to denote the existence of an EAS $a$ or $b$ in the knowledge graph $\mathcal{G}$.



An EAS $a=r(X,t)$ is said to \emph{ground} on the entity $s$, if $X=s$. All the EASs (ending anchored structures) that ground on $s$ in a given knowledge graph $\mathcal{G}$ make up the \emph{neighbourhood} of $s$~\cite{A2N}. Similarly, we can also say that the relation $r$ ($r\in \mathcal{R}\cup \mathcal{R}^{-1}$) grounds on $s$, if $r(s,t)$ is a known triple for some entity $t$. 



Then, \emph{knowledge graph reasoning} means that given the known triples in $\mathcal{T}\cup\mathcal{T}^{-1}$, we need to answer how confident we are on the correctness of a (known or unknown) triple $r(s,t)$. 
Note that reasoning on known triples is called \emph{knowledge graph cleaning}~\cite{KG_cleaning}, and reasoning on unknown ones is called \emph{knowledge graph completion}~\cite{Socher_KG_reasoning}.


\begin{figure*}[htbp]
\centering
\includegraphics[width=15cm]{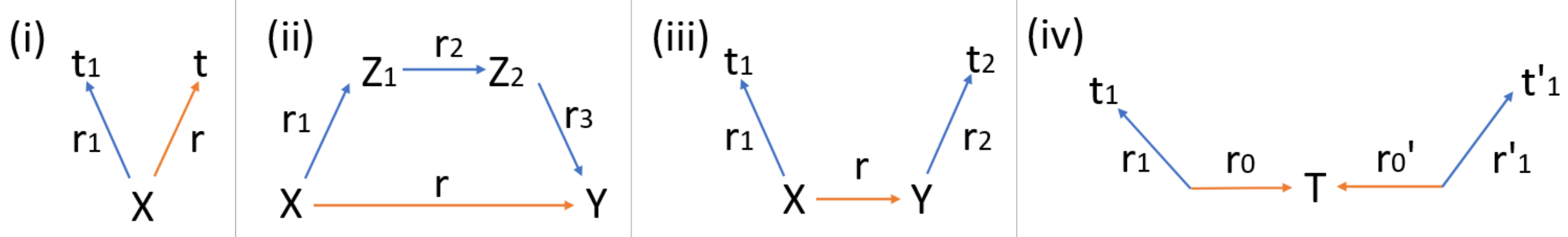}
\caption{Rule types involved in our knowledge graph reasoning model.}
\label{fig_1}
\end{figure*} 


Then in a nutshell, our theory describes the interactions among paths and ending anchored structures. We start by introducing our basic hypotheses.


\subsection{Basic Hypotheses}


Given an inverse completed knowledge graph $\mathcal{G}=(\mathcal{E},\mathcal{R}\cup\mathcal{R}^{-1},\mathcal{T}\cup\mathcal{T}^{-1})$, we have two basic hypotheses:\\

\textbf{(i) Stochastic Distribution Hypothesis}: Any relation $r$ will occur between two entities $s$ and $t$ with a probability $P(r(s,t))$.\\

\textbf{(ii) Fair Assumption Hypothesis:} Any assumption on $P(r(s,t))$, denoted as $\hat{P}(r(s,t))$, must satisfy the Principle of Maximum Entropy \cite{max_entropy} and coincide with the observations.\\

The first hypothesis indicates that a knowledge graph $\mathcal{G}$ is generated in a stochastic process (although it may not be strictly random): Moreover, if we assume that $r(X,t)$ will always occur in $\mathcal{G}$ (or $r$ will ground on $t$ with probability 1), we can directly deal with the EAS $a=r(X,t)$: We can assume that $a$ will ground on an entity $s$ with probability $P_a(s)$, which forms the probability distribution $\mathbf{P}_a$ on $\mathcal{E}$.

To understand the second hypothesis, suppose $a=r(X,t)$ grounds on $n$ entities in $\mathcal{E}$, with $|\mathcal{E}|=N$. If we know nothing else, we can only assume that $a$ will ground on any entity $s\in\mathcal{E}$ with probability $\hat{P}_a(s)=n/N$. That is, according to the Principle of Maximum Entropy, our assumption has to be ``fairly'' open to all unknown possibilities~\cite{max_entropy_2}. Also, our assumption has to be ``fairly'' based on the observed groundings.

Based on these two hypotheses, we shall introduce our knowledge graph reasoning method in the next subsection.

\subsection{Reasoning Method}
Our knowledge graph reasoning method includes four different rule types, as indicated in Figure~\ref{fig_1}: (i) ending anchored rule (EAR), (ii) cyclic anchored rule (CAR), (iii) bi-side ending anchored rule (bisEAR), and (iv) the ``rule of rule" (RofR). We shall present their reasoning processes in the following parts.

\subsubsection{Ending Anchored Rule (EAR)}

As mentioned, we assume that the EAS $a=r(X,t)$ will always occur in $\mathcal{G}$, with a probability distribution $\mathbf{P}_a$ to ground on $\mathcal{E}$. We denote our initial assumption on $\mathbf{P}_a$ to be $\mathbf{\hat{P}}_a$. 

Suppose we observe the grounding of $a$ on $n$ entities: $G_a=\{s_1,\cdots,s_n\}$. Again as mentioned, we have $\hat{P}_a(s)=n/N$ for every entity $s\in \mathcal{E}$ according to the Fair Assumption Hypothesis. To evaluate the validity of $\mathbf{\hat{P}}_a$, we choose another EAS $b=r_1(X,t_1)$ that exists in $\mathcal{G}$. Suppose $b$ grounds on $m$ entities: $G_b =\{s'_1,\cdots,s'_m\}$, 
and suppose the grounding intersection between $a$ and $b$ is $G_{a\cap b}=G_a\cap G_b = \{s_1,\cdots,s_k\}$. That is, $k$ entities in $G_b$ are grounded by $a$, and the other $m-k$ entities are not. It is easy to see that $0 \leq k \leq \min(m,n)$. 

If $\mathbf{\hat{P}}_a$ is valid, the probability for this grounding intersection situation to happen will be:
$$P(|G_{a\cap b}|=k)={m \choose k}\cdot\left(\frac{n}{N}\right)^k\cdot \left(1-\frac{n}{N}\right)^{m-k},$$
where ${m \choose k}$ means the number of $k$-combinations from $m$ elements. That is, according to $\mathbf{\hat{P}}_a$, the grounding intersection probability $P(|G_{a\cap b}|=k)$ equals to the binomial distribution probability $P(k;m,\frac{n}{N})$ in the binomial distribution $B(m,\frac{n}{N})$~\cite{binomial_distribution}.

Then, we obtain the 95$\%$ confidence interval of $B(m,\frac{n}{N})$~\cite{binomial_distribution}, denoted as $[k_0,k_1]$. 
A detailed introduction on how to generate $[k_0,k_1]$ is provided in Appendix A.



Hence, if $k < k_0$ or $k > k_1$, we will know that the probability $P(k;m,\frac{n}{N})$ (and hence $P(|G_{a\cap b}|=k)$) is too small to support our initial assumption $\mathbf{\hat{P}}_a$. If $k<k_0$, we say that $b$ \textbf{repels} $a$ on $\mathcal{E}$. If $k>k_1$, we say that $b$ \textbf{promotes} $a$ on $\mathcal{E}$. 

Therefore, we shall deny $\mathbf{\hat{P}}_a$ and make a new assumption. Again, our new assumption needs to obey the Principle of Maximum Entropy and coincide with the observations: If $r_1(X,t_1)$ grounds on any entity $s$, then $r(X,t)$ will ground on $s$ with probability $k/m$. This can be represented as: 

$$\frac{k}{m}: \ r(X,t) \longleftarrow r_1(X,t_1),$$
which is an \emph{ending anchored rule}, and can be simplified as $\mathbf{\hat{P}}_{a|b}=\frac{k}{m}$ or $a\longleftarrow b$.

On the other hand, if $s$ is not grounded by $b=r_1(X,t_1)$, we can only say that we do not observe the grounding of $b$ on $s$. Without considering other rules, the probability of $b$ grounding on $s$ will be $\hat{P}_b(s)=|G_b|/N=m/N$. Also, we observe that $n-k$ entities outside $G_b$ (denoted as $\mathcal{E}\backslash G_b$, $|\mathcal{E}\backslash G_b|=N-m$) are grounded by $a=r(X,t)$. Hence, if $s$ is not grounded by $b$, the probability for $s$ to be grounded by $a$ will be
\begin{align*}
&\hat{P}_{a|b}(s)\cdot \hat{P}_b(s)+\hat{P}_{a|\neg b}(s)\cdot \hat{P}_{\neg b}(s)\\
&=\frac{k}{m}\cdot \frac{m}{N} + \frac{n-k}{N-m}\cdot \frac{N-m}{N}=\frac{n}{N},
\end{align*}
where $\neg b$ means not being grounded by $b$. So, we have $G_{\neg b}=\mathcal{E}\backslash G_b$.

That is, if we do not observe the grounding of $b$ on an entity $s$, the rule $a\longleftarrow b$ will be irrelevant, and the initial assumption $\mathbf{\hat{P}}_a$ still applies.

\subsubsection{Cyclic Anchored Rule (CAR)}


Without assuming the definite occurrence of an ending anchored structure $r(X,t)$, we shall in general assume that $r$ will occur between two entities $s$ and $t$ with a probability $P(r(s,t))$. 

Suppose we observe $n$ known triples with the relation $r$: $\{r(s_1,t_1),\cdots,r(s_n,t_n)\}$. That is, $n$ pairs of entities $G_r = \{(s_1,t_1),\cdots,(s_n,t_n)\}$ are connected by $r$. Again, according to the Fair Assumption Hypothesis, we have to assume that $P(r(s,t))=\frac{n}{N^2}$ for any entities $s$ and $t$, where $N=|\mathcal{E}|$ is the total number of entities. That is, $r$ will occur randomly on $\mathcal{E}\times\mathcal{E}$ with probability $\frac{n}{N^2}$.

In order to estimate this initial assumption, denoted as $\mathbf{\hat{P}}_r$, we choose a path $p=r_1\!\land\! r_2\!\land\!\cdots\!\land\!r_l$. Suppose there are $m$ pairs of entities $G_p = \{(s_1,t_1),\cdots,(s_m,t_m)\}$ being connected by $p$. Then, suppose the grounding intersection between $r$ and $p$ are $k$ pairs of entities: $G_{r\cap p}=G_r\cap G_p=\{(s_1,t_1),\cdots,(s_k,t_k)\}$. 

Similarly, if the path $p$ and the relation $r$ occur independently, the probability for this grounding intersection situation to happen will be
$$P(|G_{r\cap p}|=k)={m \choose k}\cdot\left(\frac{n}{N^2}\right)^k\cdot \left(1-\frac{n}{N^2}\right)^{m-k},$$
which is the binomial distribution probability $P(k;m,\frac{n}{N^2})$ in $B(m,\frac{n}{N^2})$.

Suppose the 95$\%$ confidence interval of $B(m,\frac{n}{N^2})$ is $[k_0,k_1]$. If $k$ falls out of $[k_0,k_1]$, we shall deny $\mathbf{\hat{P}}_r$ and make a new assumption referring to the path $p$ on $\mathcal{E}\times \mathcal{E}$. This leads to the rule:
$$r(X,Y)\!\longleftarrow\! r_1(X,\!Z_1\!)\!\land\! r_2(Z_1,\!Z_2\!)\!\land\!\cdots\!\land\!r_l(Z_{l-1},Y),$$
which is a \emph{cyclic anchored rule}, or a \emph{Horn clause rule} \cite{DRUM}. We can further simplify it as $\mathbf{\hat{P}}_{r|p}=\frac{k}{m}$ or $r\longleftarrow p$. 

Similar to the situation in ending anchored rule, if path $p$ does not occur between entities $s$ and $t$, the initial assumption $\mathbf{\hat{P}}_r$ still applies. To avoid redundancy, we will not provide the analysis again. We note that the EAR and CAR are typical rules used in many KG completion models \cite{DRUM,NeuralLP}. In the followings subsections, we shall introduce two novel rule types.


\subsubsection{Bi-side Ending Anchored Rule (bisEAR)}

Now, we shall introduce a novel rule type, whose structure is roughly shown in Figure \ref{fig_1} (iii). 

Again, given a relation $r$, we make our initial assumption $\mathbf{\hat{P}}_r$ as: $r$ occurs between any two entities $s$ and $t$ with a probability $\frac{n}{N^2}$, where $n$ is the number of known triples with relation $r$ and $N$ is the total number of entities. 

But then, instead of referring to paths to evaluate $\mathbf{\hat{P}}_r$, we look at the ending anchored structures (EASs) that could ground on both the $X$-side and the $Y$-side of the triple template $r(X,Y)$: Suppose $b_1 = r_1(X,t_1)$ grounds on $m_1$ entities $S=\{s_1,\cdots,s_{m_1}\}$, and $b_2 = r_2(Y,t_2)$ grounds on $m_2$ entities $T=\{t_1,\cdots,t_{m_2}\}$. Then, suppose there are $k$ pairs of entities within $S\times T$ being connected by $r$. That is, we suppose that $r(s_{i_1},t_{i_1}),\cdots,r(s_{i_k},t_{i_k})$ are known triples, with $s_{i_1},\cdots,s_{i_k}\in S$ and $t_{i_1},\cdots,t_{i_k}\in T$.

If the occurrence of $r$ between two entities $s$ and $t$ are independent with respect to $b_1$ and $b_2$, the probability of the above event to happen will be:
$$P={m_1\cdot m_2 \choose k}\cdot\left(\frac{n}{N^2}\right)^k\cdot \left(1-\frac{n}{N^2}\right)^{m_1\cdot m_2-k},$$
which is the binomial probability $P(k;m_1m_2,\frac{n}{N^2})$ in the binomial distribution $B(m_1m_2,\frac{n}{N^2})$.

Similarly, suppose the 95$\%$ confidence interval of $B(m_1m_2,\frac{n}{N^2})$ is $[k_0,k_1]$. We will reject $\mathbf{\hat{P}}_r$ if $k$ falls out of $[k_0,k_1]$, which leads to the rule 
$$\frac{k}{m_1\cdot m_2}: \ r(X,Y)\longleftarrow r_1(X,t_1) \ \& \  r_2(Y,t_2).$$

We call it a \emph{bi-side ending anchored rule} (bisEAR) and can be simplified as $\mathbf{\hat{P}}_{r|b_1\&b_2}=\frac{k}{m_1m_2}$ or $r\longleftarrow b_1\&b_2$. Note that we use the symbol ``$\&$" to denote the combination of $b_1$ and $b_2$, which is different from the meaning of the ordered chain $r_1\land r_2$. And again, if one of the grounding conditions ($b_1$ and $b_2$) is not fulfilled, the initial assumption $\mathbf{\hat{P}}_r$ still applies.

\subsubsection{The Rule of Rule (RofR)}
Finally, we introduce another novel rule type, whose frame is roughly shown in Figure \ref{fig_1} (iv).

Suppose we learned an ending anchored rule (EAR): $r_0(X,t_0)\longleftarrow r_1(X,t_1)$. Then, it is obvious that we can re-write it in the form: $(r_0\circ r_1)(t_0,t_1)$. If we define a new relation as $r_{0,1}=(r_0\circ r_1)$, we can get a new triple: $r_{0,1}(t_0,t_1)$.

Then, writing all the learned ending anchored rules (EARs) into such a form, we can get the new ``rule triples". After that, we will obtain a new knowledge graph $\tilde{\mathcal{G}}$ generated by the rule triples. Finally, we can apply the same reasoning process on $\tilde{\mathcal{G}}$ to get its corresponding rules in different types: EAR, CAR and bisEAR with respect to $\tilde{\mathcal{G}}$, which we call the \emph{rule of rule} (RofR). 

Due to the computational complexity, we only apply the ending anchored rule (EAR) on $\tilde{\mathcal{G}}$ in this paper. The reasoning process is exactly the same as on $\mathcal{G}$: Given a \emph{rule ending anchored structure} (REAS) $\tilde{a}=r_{0,1}(T,t_1)$, suppose it grounds on entities $G_{\tilde{a}}=\{t_{0_1},\cdots,t_{0_n}\}$, which leading to our initial assumption $\mathbf{\hat{P}}_{\tilde{a}}=\frac{n}{N}$ on the entire $\mathcal{E}$. In order to evaluate $\mathbf{\hat{P}}_{\tilde{a}}$, we use another REAS $\tilde{b}=r_{0,1}'(T,t_1')$ to obtain the grounding intersection $G_{\tilde{a}\cap \tilde{b}}=G_{\tilde{a}}\cap G_{\tilde{b}}$, where $G_{\tilde{b}}$ is the groundings of $\tilde{b}$.

Suppose $|G_{\tilde{b}}|=m$ and $|G_{\tilde{a}}\cap G_{\tilde{b}}|=k$. Again, we have that $P(|G_{\tilde{a}\cap \tilde{b}}|=k)=P(k;m,\frac{n}{N})$, where $P(k;m,\frac{n}{N})$ is the binomial probability in the binomial distribution $B(m,\frac{n}{N})$. 

And again, by checking whether $k$ falls out of the 95$\%$ confidence interval $[k_0,k_1]$, we will decide whether we reject $\mathbf{\hat{P}}_{\tilde{a}}$. But different from previous rule types, we will make a new assumption $\mathbf{\hat{P}}_{\tilde{a}|\tilde{b}}=\frac{k}{m}$ only if $k> k_1$. This is because repelling from $\tilde{b}$ to $\tilde{a}$ cannot provide useful ending anchored rules, which will be further explained shortly. This will be the \emph{rule of ending anchored rule} (REAR), which can be represented as
$$\frac{k}{m}: \ r_{0,1}(T,t_1)\longleftarrow r_{0,1}'(T,t_1'),$$
or be simplified as $\tilde{a}\longleftarrow\tilde{b}$. Figure \ref{fig_1} (iv) shows its frame.

Now, with the REAR $\tilde{a}\longleftarrow\tilde{b}$ on $\tilde{\mathcal{G}}$, we can obtain new EAR on $\mathcal{G}$: Given an entity $t\in G_{\tilde{b}}$, we have that $\tilde{a}=r_{0,1}(T,t_1)$ will ground on $t$ with probability $k/m$ according to $\mathbf{\hat{P}}_{\tilde{a}|\tilde{b}}=\frac{k}{m}$. This means that the probability for $r_{0,1}(t,t_1)$ to be a valid rule triple is $k/m$. In other words, the probability for $r_0(X,t)\longleftarrow r_1(X,t_1)$ to be a valid ending anchored rule (EAR) is $k/m$. 

If we have already obtained $r_0(X,t)\longleftarrow r_1(X,t_1)$ from the EAR reasoning process on $\mathcal{G}$, it is redundant to have another exactly the same rule with different probability. So, $r_0(X,t)\longleftarrow r_1(X,t_1)$ is valuable only when it is a new rule, which cannot be directly obtained from the EAR reasoning process on $\mathcal{G}$. But in that case, we need to decide the probability of $r_0(X,t)\longleftarrow r_1(X,t_1)$.

Going back to the REAR $\tilde{a}\longleftarrow\tilde{b}$ (recall that $\tilde{a}=r_{0,1}(T,t_1)$ and $\tilde{b}=r_{0,1}'(T,t_1')$): Suppose the REAS grounding intersection is $G_{\tilde{a}\cap\tilde{b}}=\{t_{0_1},\cdots,t_{0_k}\}$. Then, from the way we create REAS, we can see that $r_{0,1}(t_{0_i},t_1)$ are all known rule triples for $i=1,\cdots,k$. That is, $r_0(X,t_{0_i})\longleftarrow r_1(X,t_1)$ are all learned EAR based on the initial knowledge graph $\mathcal{G}$, for $i=1,\cdots,k$. Suppose their probabilities distributions are $\mathbf{\hat{P}}_{a_i|b}=P_i$ for $i=1,\cdots,k$, with $a_i = r_0(X,t_{0_i})$ and $b = r_1(X,t_1)$. Then, we take the average probability $\bar{P}=(\sum_{i=1}^kP_i)/k$, which will be an estimation on the probability of $r_0(X,t)\longleftarrow r_1(X,t_1)$. Finally, we cannot ignore the probability $\mathbf{\hat{P}}_{\tilde{a}|\tilde{b}}=k/m$, indicating that how confidant we are on $r_0(X,t)\longleftarrow r_1(X,t_1)$ to be a rule. Hence, the final \emph{estimated ending anchored rule} (EEAR) will be:
$$\alpha\cdot\hat{P}_{\tilde{a}|\tilde{b}}\cdot \bar{P}: \ r_0(X,t)\longleftarrow r_1(X,t_1),$$
where $\alpha$ is an empirically chosen weight for the RofR rule type. We always set $\alpha=0.2$ in this paper. Similarly, the above EEAR can be simplified as $\mathbf{\hat{P}}_{a|b}=\alpha\cdot\hat{P}_{\tilde{a}|\tilde{b}}\cdot \bar{P}$ or $a\longleftarrow b$, with $a=r_0(X,t)$. Since the repelling from $\tilde{b}$ to $\tilde{a}$ will make $\hat{P}_{\tilde{a}|\tilde{b}}$ (and hence $\hat{P}_{\tilde{a}|\tilde{b}}\cdot \bar{P}$) near zero, we do not consider the situation of $\tilde{b}$ repels $\tilde{a}$.

After obtaining the rules from all four different types, we shall apply them on each triple $r(s,t)$, which is introduced in the next section.

\section{Implementations and Performances}
\label{sec:imp_perf}
In this section, we will implement our theory by what we call the \emph{RuleDict} model, which will be tested on three large benchmark datasets of knowledge graph completion. Each of the four rule types is implemented in its specific way, which will be introduced accordingly.

\subsection{Implementation of EAR}
\label{ssec:imp_EAR}

Suppose we have an inverse completed knowledge graph $\mathcal{G}=(\mathcal{E},\mathcal{R}\cup\mathcal{R}^{-1},\mathcal{T}\cup\mathcal{T}^{-1})$, with $|\mathcal{R}|=R$ and $|\mathcal{E}|=N$. We first build the grounding set $G_a$ for each $a=r(X,t)\in\mathcal{G}$: $G_a=\{s|s\in\mathcal{E}, r(s,t)\in\mathcal{T} \cup \mathcal{T}^{-1}\}$. 

Then, we create the $\emph{connection set}$ $C_a$ for each $a=r(X,t)\in\mathcal{G}$: $C_a=\{b|b=r_1(X,t_1)\in\mathcal{G}, b\neq a, |G_{a\cap b}|\neq \emptyset \}$, where $G_{a\cap b}=G_a\cap G_{b}$. That is, the connection set $C_a$ stores all the other EASs $b$ that have grounding intersections with $G_a$ (``connected" to $a$). 




\begin{algorithm}
\SetAlgoLined
 \For{$a=r(X,t)\in \mathcal{G}$}{\For{$b\in C_a$}{ \If{$b$ promotes \textbf{or} repels $a$ on $\mathcal{E}$}{$\mathbf{\hat{P}}_{a|b}$: $\hat{P}_a(s)=\frac{|G_{a\cap b}|}{|G_{b}|}$, $s\in G_b$\;
 \emph{Store} $\mathbf{\hat{P}}_{a|b}: a\longleftarrow b$ w.r.t $a$ }\textbf{end}}...}\textbf{end}
\caption{Implementation of EAR}
\label{alg_1}
\end{algorithm}

Then, we evaluate the initial assumption $\mathbf{\hat{P}}_a$ using each $b\in C_a$ as described in Section 2.3.1. 
Algorithm \ref{alg_1} shows the process.

\subsection{Implementation of CAR}
\label{ssec:imp_CAR}


Before we start, for each $r\in\mathcal{R}$, we shall obtain all the test triples with relation $r$: $\{r(s_1,t_1),\cdots,r(s_{n_r},t_{n_r})\}\subseteq \mathcal{G}_{test}$. Then, the known source entities in these triples form the \emph{source set} for $r$: $S_r=\{s_1,\cdots,s_{n_r}\}\cap\mathcal{E}$, and similarly we can get the \emph{target set} $T_r=\{t_1,\cdots,t_{n_r}\}\cap\mathcal{E}$. That is, we do not consider new entities from the test triples. 

Given a path $p=r_1\land \cdots \land r_l$, we can see that its source set (formed by training triples) is $G_{r_1}$, the groundings of $r_1$. Its target set is $G_{r_l^{-1}}$, the groundings of inverse $r_l$. Note that $p$ is an existing path in $\mathcal{G}$ (denoted as $p\in\mathcal{G}$) only if $G_{r_i^{-1}}\cap G_{r_{i+1}}\neq \emptyset$ for $i=1,\cdots,l-1$. 

Now, we will observe each $p=r_1\land \cdots \land r_l$ with length $l=1,2,3$ to discover cyclic anchored rules (CAR). We will skip a path $p$ if it does not exist in $\mathcal{G}$. For each qualified path $p$, we will consider its potential CAR with respect to each $r\in\mathcal{R}$, as introduced in Section 2.3.2. 
This is shown in algorithm \ref{alg_2}.

\begin{algorithm}
\SetAlgoLined
 \For{$p=r_1\land \cdots \land r_l\in\mathcal{G}$ with $l\leq 3$}{\For{$r\in \mathcal{R}$ with $\big($ $G_{r_1}\cap S_r\neq\emptyset$ \textbf{or} $G_{r_l^{-1}}\cap T_r\neq \emptyset \big)$}{ \If{$p$ promotes \textbf{or} repels $r$ on $\mathcal{E}\times \mathcal{E}$}{$\mathbf{\hat{P}}_{r|p}$: $\hat{P}_r(s,t)=\frac{|G_{r\cap p}|}{|G_{p}|}$, $(s,t)\in G_r$\;
 \emph{Store} $\mathbf{\hat{P}}_{r|p}: r\longleftarrow p$ w.r.t $r$ }\textbf{end}}...}\textbf{end}
\caption{Implementation of CAR}
\label{alg_2}
\end{algorithm}

Here, the condition $G_{r_1}\cap S_r\neq\emptyset$ \textbf{or} $G_{r_l^{-1}}\cap T_r\neq \emptyset \big)$ is to guarantee that the rule $r\longleftarrow p$ is useful for ranking the test triples in the source and target prediction, which will be introduced in the evaluation protocol shortly. That is, our rule learning is \emph{task-specific}: We do not learn rules useless for our task.


\begin{table*}[htbp]
\begin{center}
\resizebox{.8\textwidth}{!}{
\begin{tabular}{l|cccc|cccc}
\hline
\multirow{2}{*}{\textbf{Model}}&\multicolumn{4}{c}{\textbf{WN18RR}} & \multicolumn{4}{c}{\textbf{FB15K-237}}\\
\cline{2-9}
&@$1$&@$3$&@$10$&MRR&@$1$&@$3$&@$10$&MRR\\
\hline
DRUM~\cite{DRUM}&42.5&\textbf{51.3}&\textbf{58.6}&\underline{0.486}&25.5&37.8&51.6&0.343\\
RotatE~\cite{RotatE}&42.8&49.2&57.1&0.476&24.1&37.5&53.3&0.338\\
SACN~\cite{SACN}&43.0&48.0&54.0&0.470&\underline{26.0}&\underline{39.0}&\underline{54.0}&\underline{0.350}\\
MuRP~\cite{MuRP}&44.0&49.5&56.6&0.481&24.3&36.7&51.8&0.335\\
A2N~\cite{A2N}&42.0&46.0&51.0&0.450&23.2&34.8&48.6&0.317\\
GPFL~\cite{GPFL}&\underline{44.9}&\underline{50.0}&55.2&0.480&24.7&36.2&50.4&0.322\\
ConvE v6~\cite{ConvE}$\!\!\!$&40.0&44.0&52.0&0.430&23.7&35.6&50.1&0.325\\
ComplEx-N3~\cite{complex_N3}$\!\!$&43.5&49.5&\underline{57.2}&0.480&\textbf{26.4}&\textbf{39.2}&\textbf{54.7}&\textbf{0.357}\\
ComplEx~\cite{ComplEx}$\!\!\!$&41.0&46.0&51.0&0.440&15.8&27.5&42.8&0.247\\
DisMult~\cite{DisMult}&39.0&44.0&49.0&0.430&15.5&26.3&41.9&0.241\\
\hline
RuleDict&\textbf{45.0}&\underline{50.0}&55.9&\textbf{0.487}&22.6&32.9&46.5&0.305\\
\hline
\end{tabular}}
\end{center}
\caption{Experimental results on WN18RR and FB15K-237 test sets. Hits@N scores are in percentage. The best score is in \textbf{bold} and the second best one is \underline{underlined}.}
\label{tab_1}
\end{table*}

\begin{figure*}[htbp]
\centering
\includegraphics[width=15cm]{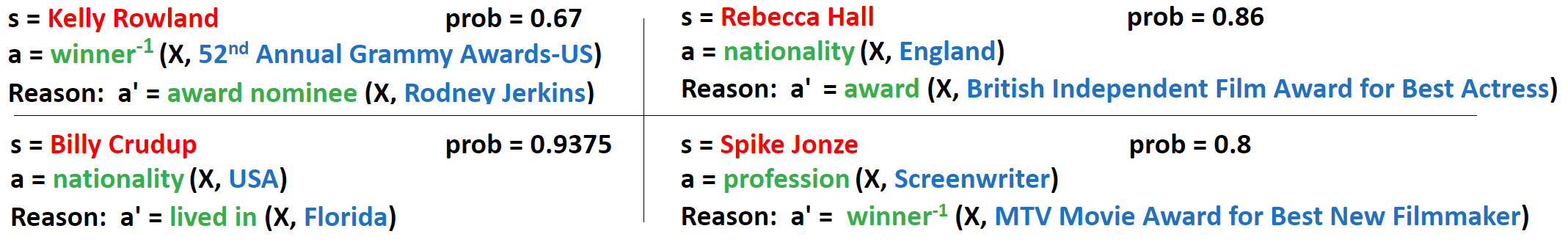}
\caption{The probability and reason for each EAS $a=r(X,t)$ to ground on an entity $s$. Here, \emph{winner} is short for \emph{/award/award$\_$ceremony/} \emph{awards$\_$presented./award/award$\_$honor/award$\_$winner}. Then, \emph{award nominee} is short for \emph{/award/award$\_$nominee/award$\_$nominations./award/award$\_$} \emph{nomination/award$\_$nominee}. And then \emph{award} is short for \emph{/award/award$\_$nominee/award$\_$nominations./award/award$\_$nomination/award}. Other relations are easy to understand and locate in the FB15K-237 dataset.}
\label{fig_2}
\end{figure*}

\subsection{Implementation of bisEAR and RofR}
\label{ssec:imp_RofR}

We will also apply task-specific implementation for the bisEAR and RofR rule types, which are introduced in Appendix B and C.

Note that all these rules are stored by Python nested dictionaries \cite{Python_intro}. Therefore, we call our model the \emph{RuleDict} model.

\subsection{Applying the rules}
\label{ssec:apply_rules}

For each triple $r(s,t)$, we will refer to all the rules that can be applied: We will evaluate if $r(X,t)$ grounds on $s$ for some reason $r_1(X,t_1)$, or $r^{-1}(Y,s)$ grounds on $t$ for some reason $r_2(Y,t_2)$, or $r$ will connect $(s,t)$ in a CAR or bisEAR type of rule. Then, we get the corresponding prediction probability scores from each applied rule. We sort all these scores in a descending order. For each triple, we record at most 10 probability scores. We full-fill with zeros if there are not enough scores.

Then, we apply the routine \emph{filtered} evaluation protocol~\cite{KG_evaluation_protocol,WN18RR}. That is, given a test triple $r(s,t)$, we want to rank it among all the $r(s,t')$ in a descending order by the first probability score of each triple. If there is a tie in the first probability score, we will refer to the second one, and so on. Here, $t'\in\mathcal{E}$ and $r(s,t')$ is not in any of the training, validation and test set. This is called target prediction. Similarly, we want to rank $r(s,t)$ among all the $r(s',t)$ in the source prediction, with each $r(s',t)$ to be filtered as well. Then, averaged metrics from both predictions are reported. We report mean reciprocal rank (MRR) and the proportion of correct entities in the top $N$ ranks (Hits@$N$) for $N=$ 1, 3 and 10.

Note that some knowledge graph completion tasks, such as the FB15K-237, remove a validation or test triple $r(s,t)$ if the entity pair $(s,t)$ is connected in the training set \cite{FB15K_237}. To deal with this issue, we obtain all the training entity pairs (suppose there are $n$ of them) and all the validation entity pairs (suppose there are $m$ of them). Then, we obtain the repeated pairs. Suppose there are $k$ of them. 

Again, we assume the training and validation entity pairs occur independently in $\mathcal{E}\times\mathcal{E}$, which leads to the binomial probability $P(k;m,\frac{n}{N^2})$ with $|\mathcal{E}|=N$. And again, suppose $[k_0,k_1]$ is the 95$\%$ confidence interval of $B(m,\frac{n}{N^2})$. If $k<k_0$, we will know that the repeated triples are removed, which we assume will also be applied to the test triples. As a result, for a test triple $r(s,t)$, if $(s,t)$ is connected in the training set, we will only append zeros to its probability score list.

\subsection{Performances}
We implement our RuleDict model on the knowledge graph completion tasks WN18RR~\cite{WN18RR} and FB15K-237~\cite{FB15K_237}, which are the subset datasets of WN18 and FB15K respectively, with the trivial reversible relations removed~\cite{WN18RR}. Also, we discovered that FB15K-237 removes the occurred entity pairs in the test and validation set as introduced in the previous section, yet WN18RR does not. As a result, for a triple with occurred entity pair in FB15K-237, we will only append zeros to its probability score list. Using the machine with Intel Xeon(R) CPU E5-2637 v3 @ 3.50GHz $\times$ 16 cores and 128 GiB memory, it takes approximately 3 hours to run the RuleDict model on WN18RR, and 30 hours on FB15K-237.


We compare the performances of our models with those of several benchmark models. The results are shown in Table \ref{tab_1}. All the results are taken from their original papers. We can see that our model achieves several state-of-the-art results on WN18RR, and is competitive on FB15K-237. 

Moreover, our models also possess advantages beyond these scored performances. We choose some EAR $a\longleftarrow a'$ with $a=r(X,t)$ and $a'=r'(X,t')$ in FB15K-237. Then we choose some entities in the groundings $G_{a'}$ to show that $a$ grounds on $s$ with $a'$ as the reason. The results are shown in Figure~\ref{fig_2}.

That is, as mentioned in the previous section, our model can provide the precise reason explaining the occurrence of a triple $r(s,t)$. The grounding of relation $r$ on a pair of entity $(s,t)$ with respect to bisEAR or CAR can be exhibited in a similar way, which will not be repeated here. A further discussion of the examples in Figure \ref{fig_2} is given in the next section.

\section{Discussion on some EAR examples}
\label{sec:disc}

In this section, we shall analyze the first example in Figure \ref{fig_2} of the main paper, which is provided again here. 

According to our theory, any example in Figure \ref{fig_2} shall be interpreted as: The probability for $a=r(X,t)$ to ground on $s$ is $p$, if $s$ is grounded by $a'=r'(X,t')$. 

Then, the first example indicates that, the probability for Kelly Rowland to win the 52nd US Grammy Awards is 0.67. This is because Kelly is an award nominee (co-nominee) of Rodney Jerkins, and the probability for Rodney's co-nominee to win the 52nd US Grammy Award is 0.67.

To further evaluate our model, we look deeper into this example: According to the known triples in FB15K-237, there are six entities (artists) who are co-nominees of Rodney Jerkins: Mark Stent, Lady Gaga, RedOne, Jay-Z, Brandy Norwood and Kelly Rowland. Four of them (except Brandy Norwood and Kelly Rowland) won the 52nd US Grammy Award. And again, according to the known triples, there are in total 54 artists who won the 52nd US Grammy Award. 

Initially, we assume that any entity in FB15K-237 will randomly win the 52nd US Grammy Award with probability $54/14541\approx 0.00371$ ($|\mathcal{E}|=14541$ in FB15K-237). Then, the 95$\%$ confidence interval of $B(6,0.00371)$ is $[0,0]$. That is, the initial assumption shall be denied, even if only one of the above six artists wins the 52nd US Grammy Award. But now, four of them won this award, which definitely falls out of the confidence interval. As a result, our model obtains the corresponding EAR: 
\begin{align*}
&P=\frac{4}{6}: \texttt{win 52nd US Grammy Award}\\
&\longleftarrow \texttt{co-nominee of Rodney Jerkins}. 
\end{align*}


Therefore, although Kelly Rowland did not win the 52nd US Grammy Award (according to the FB15K-237 dataset), we still regard the first example in Figure \ref{fig_2} to be reasonable with an estimated probability of $\frac{4}{6} \approx 0.67$.

\section{Related Work}
\label{sec:related_work}
The A2N model~\cite{A2N} is the closest work to ours, in which entity-relation compound embeddings are learned, which are then used to predict the missing entity via an attention scoring process. The A2N model provides reasons behind their predictions via this attention scoring process. 

However, due to the scoring nature of attention model~\cite{AttentionIsAll}, reasons provided by the A2N model are only linearly weighted by the attention scores, which are not precise. In contrast, our models can provide reasons with precise probabilities for each evaluated triple.

Then, the GPFL model introduced in~\cite{GPFL} is another model that is closely related to ours. The GPFL model is also built on ending anchored rules (EARs), which are referred to as both anchored rules (BARs) in their paper. However, the GPFL model does not provide a fundamental theory for knowledge graph reasoning, and our RuleDict model also outperforms the GPFL model on the benchmark knowledge graph completion task WN18RR.

Beyond, there are other rule-based knowledge graph completion models, such as NeuralLP~\cite{NeuralLP} and MINERVA~\cite{MINERVA}. The NeuralLP algorithm generates differentiable models for learning logical rules. The MINERVA model applies reinforcement learning~\cite{reinforcement_learning} with random walk in the knowledge graph reasoning process. 

\section{Conclusion and Future Plans}
\label{sec:conclusion}
In this paper, we present a 
theory for knowledge graph reasoning, which can provide a precise reason explaining why a triple is believed to be correct. Then, we implement our theory by the RuleDict model, which outperforms other models on the knowledge graph completion task WN18RR and provides competitive results on FB15K-237. 

We admit that rule-based models usually lack of scalability, which cannot be applied to industrial level knowledge graphs. To overcome this issue, we hope to design the class-rule-based knowledge graph reasoning model in the future. That is, we classify entities in an unsupervised manner, such that all the discovered rules (CAR, EAR, bisEAR) can be explained by the relation connection between entity classes. In such a way, we hope to keep both the scalability and the reasoning capacity of the rule-based model.

\bibliography{custom}
\bibliographystyle{IEEEtran}

\appendix

\section{Binomial distribution and its approximation}
\textbf{Part A}: In this appendix, we shall introduce how to obtain the 95$\%$ confidence interval of the binomial distribution $B(m,\frac{n}{N})$ mentioned in Section 2.3.1.

\begin{figure}[htbp]
\centering
\includegraphics[width=7.5cm]{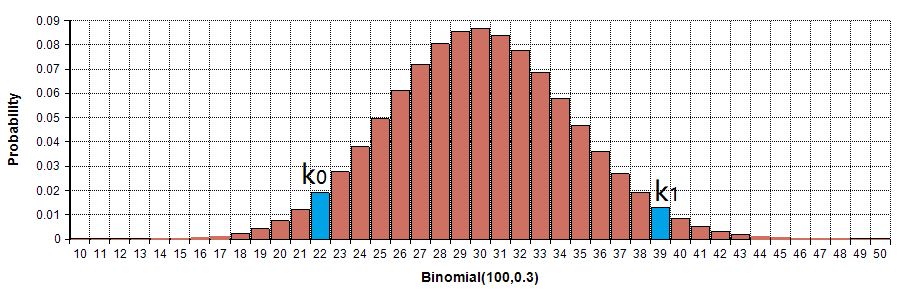}
\caption{The 95$\%$ confidence interval of $B(100,0.3)$.}
\label{fig_3}
\end{figure}

According to the property of the binomial distribution, $P(j;m,\frac{n}{N})$ descends from the expected value $m\cdot\frac{n}{N}$ of $B(m,\frac{n}{N})$ towards both directions on the axis of $j$. So, gathering $j=0,1,\cdots,m$, we put the values of $P(j;m,\frac{n}{N})$ in descending order. Then, we choose the fewest leading values such that their summation $\sum_jP(j;m,\frac{n}{N}) \geq 0.95$. Suppose $k_0$ is the minimum of the chosen $\{j\}$ and $k_1$ is their maximum. Then, we can see that $[k_0,k_1]$ is the 95$\%$ confidence interval of $B(m,\frac{n}{N})$. Figure~\ref{fig_3} shows the $95\%$ confidence interval, $[22,39]$, of the binomial distribution $B(100,0.3)$. 

However, when $m$ is large, it is expensive to obtain the 95$\%$ confidence interval in the above way. So, when $m > 15$, we use the normal distribution $\mathcal{N}(\mu,\sigma^2)$ to approximate $B(m,\frac{n}{N})$, where $\mu=\frac{mn}{N}$ and $\sigma=\frac{\sqrt{mn(N-n)}}{N}$ \cite{hansen2011approximating}. Then, we have that $k_0 = \mathbf{int}(\mu-2\sigma)$ and $k_1=\mathbf{int}(\mu+2\sigma)$, where $\mathbf{int}$ means to round to the closest integer \cite{altman2005standard}. 

\section{Implementation of bisEAR}
\label{ssec:imp_bisEAR}

\textbf{Part B:} This appendix part will introduce the implementation process of bisEAR.

Similarly, the implementation of bi-side ending anchored rule (bisEAR) is also task-specific. For each $r\in\mathcal{R}$, we obtain all the training triples with relation $r$: $\{r(s_1,t_1),\cdots,r(s_{n_r},t_{n_r})\}\subseteq \mathcal{G}$. Then, we obtain the source set $\hat{S}_r$ and target set $\hat{T}_r$ with respect to these training triples.

Then, we create the \emph{source connection set} $\hat{C}_{r,S}$ as $\hat{C}_{r,S}=\{b|b=r(X,t)\in\mathcal{G},G_b\cap \hat{S}_r\neq\emptyset\}$, and the \emph{target connection set} $\hat{C}_{r,T}$ as $\hat{C}_{r,T}=\{b|b=r(X,t)\in\mathcal{G},G_b\cap \hat{T}_r\neq\emptyset\}$. Similarly, we can create the source and target connection set $C_{r,S}$ and $C_{r,T}$ for $S_r$ and $T_r$ respectively, where $\hat{S}_r$ and $\hat{T}_r$ are replaced by $S_r$ and $T_r$.

Now, given each $r\in\mathcal{R}$, for each EAS $b_1\in C_{r,T}$, we will discover its potential bisEAR with respect to each $b_2\in \hat{C}_{r,S}$ (and store $r\longleftarrow b_1\&b_2$ in the file $r_{bisEAR}.json$), which will be applied to source prediction. This is shown in Algorithm \ref{alg_3}. Accordingly, for each EAS $b_2'\in C_{r,S}$, we will discover its potential bisEAR with respect to each $b_1'\in \hat{C}_{r,T}$, which will be applied to target prediction. We do not present the algorithm due to symmetry.

\begin{algorithm}
\SetAlgoLined
 \For{$r\in\mathcal{R}$}{\For{$b_1\in C_{r,T}$}{\For{$b_2\in \hat{C}_{r,S}$}{ \If{$b_1, b_2$ promotes \textbf{or} repels $r$ on $\mathcal{E}\times \mathcal{E}$}{$\mathbf{\hat{P}}_{r|b_1,b_2}$: $\hat{P}_r(s,t)=\frac{k}{m_1m_2}$, $t\in G_{b_1}$, $s\in G_{b_2}$\;
 \emph{Store} $\mathbf{\hat{P}}_{r|b_1,b_2}: r\longleftarrow b_1,b_2$ w.r.t $r$ }\textbf{end}}...}\textbf{end}}
\caption{Implementation of bisEAR}
\label{alg_3}
\end{algorithm}

\section{Implementation of RofR}

\textbf{Part C}: In this appendix section, we shall introduce the task-specific implementation of the RofR rule type.

As mentioned in Section 2.3.4, the purpose of the RofR reasoning is to discover new rule $r_0(X,t)\longleftarrow r_1(X,t_1)$ which cannot be directly obtained from the initial knowledge graph $\mathcal{G}$. In other words, suppose we learned a rule of ending anchored rule (REAR) $\tilde{a}\longleftarrow \tilde{b}$ with $\tilde{a}=r_{0,1}(T,t_1)$ $\tilde{b}=r_{0,1}'(T,t_1')$. Then, we are interested in any entity $t\in G_{\tilde{b}}\backslash G_{\tilde{a}}=G_{\tilde{b}\backslash\tilde{a}}$. This is because then the induced rule triple $r_{0,1}(t,t_1)$ will be a new one, which is an estimated ending anchored rule (EEAR) $r_0(X,t)\longleftarrow r_1(X,t_1)$ that cannot be directly learned from $\mathcal{G}$.

Then, we suppose to learn all these new rule triples which are related to the source and target prediction for the test triples. Given a test triple $r_0(s,t)$, we take the target prediction $r_0(s,?)$ as our example: 

$(i)$ We want to discover all the new rules $r_0^{-1}(X,s)\longleftarrow r_1(X,t_1)$ from the RofR reasoning. This means that we want to discover the new rule triple $r_{0^{-1},1}(s,t_1)$ by the RofR reasoning. That is, we want to evaluate any potential REAR $r_{0^{-1},1}(X,t_1)\longleftarrow r_{2,3}(X,t_3)$ if $r_{2,3}(X,t_3)$ grounds on $s$.

Remember that given the test triples, we form the source set $S_{r_0}$ and target set $T_{r_0}$ for the relation $r_0$ as introduced in Section 3.2. Hence, given the relation $r_0$, we will first obtain the source set $S_{r_0}$. Then we obtain all the rule ending anchored structure (REAS) $r_{i,j}(X,t_j)$ that have grounding intersections with $S_{r_0}$. After that, for each of the REAS $r_{i,j}(X,t_j)$, we will observe if it has grounding intersection with any REAS of the type $r_{0^{-1},k}(X,t_k)$, with $k$ and $t_k$ to be any relation and any entity, respectively. Then, if $r_{0^{-1},k}(X,t_k)\longleftarrow r_{i,j}(X,t_j)$ is indeed a REAR, we will ground $r_{0^{-1},k}(X,t_k)$ on $S_{r_0}$ to obtain the new rule triple $r_{0^{-1},k}(s,t_k)$. Hence, we will obtain a new EEAR $r_0^{-1}(X,s)\longleftarrow r_k(X,t_k)$ as desired. This rule will be applied to the target prediction $r_0(s,?)$

$(ii)$ Also, we want to discover all the new rules $r_0(X,t')\longleftarrow r_1'(X,t_1')$ if $r_1'(X,t_1')$ grounds on $s$. This means that we want to discover the new rule triple $r_{0,1'}(t',t_1')$, if the EAS $r_1'(X,t_1')$ in the initial knowledge graph $\mathcal{G}$ grounds on $s$. That is, we want to evaluate any potential REAR $r_{0,1'}(X,t_1')\longleftarrow r_{2,3}(X,t_3)$ with any REAS $r_{2,3}(X,t_3)$.

Hence, given the relation $r_0$, we first obtain its source set $S_{r_0}$. Then, we obtain all the ending anchored structures (EAS) $r_k'(X,t_k')$ that have grounding intersection with $S_{r_0}$. Then, we will obtain all the existing REAS $r_{0,k'}(X,t_k')$ with respect to each $r_k'(X,t_k')$. For each $r_{0,k'}(X,t_k')$, we will evaluate each of its potential REAR $r_{0,k'}(X,t_k')\longleftarrow r_{i,j}(X,t_j)$ with respect to each REAS $r_{i,j}(X,t_j)$ that have a grounding intersection with this $r_{0,k'}(X,t_k')$. Suppose we indeed find one REAR $r_{0,k'}(X,t_k')\longleftarrow r_{i,j}(X,t_j)$ with the grounding of $r_{i,j}(X,t_j)$ to be $G$. Then, for each $t'\in G$, we will obtain the new test triple $r_{0,k'}(t',t_k')$, which is the new EEAR $r_0(X,t')\longleftarrow r_k'(X,t_k')$. Here, $r_k'(X,t_k')$ will have grounding intersection with $S_{r_0}$ as desired. Full-filling each $s\in S_{r_0}$ into the rule, we can obtain $r_0(s,t')\longleftarrow r_k'(s,t_k')$, which will be applied to the target prediction of $r_0(s,?)$.

The source prediction $r_0(?,t)$ can be transformed into the source prediction $r_{0^{-1}}(t,?)$, which can be handled by exactly the same task-specific RofR reasoning process.

\end{document}